\documentclass[11pt,letterpaper]{article}
\usepackage{emnlp2017}
\usepackage{times}
\usepackage{latexsym}
\usepackage{array}
\usepackage[linesnumbered,ruled]{algorithm2e}
\usepackage{multicol}
\usepackage{tabularx, booktabs}
\usepackage{graphicx}
\usepackage{amsmath}
\usepackage{color}
\usepackage{fixltx2e}
\usepackage{epstopdf}
\DeclareGraphicsExtensions{.png,.jpg}
\emnlpfinalcopy

\def\emnlppaperid{1145}

\title{Two-Stage Synthesis Networks for Transfer Learning in Machine Comprehension}  

\author{David Golub\thanks{\hspace{1mm} Work performed while interning at Microsoft Research.} \\
  Stanford University \\
  {\small \tt golubd@cs.stanford.edu} \\
  \\\And
  Po-Sen Huang \and Xiaodong He \\
  Microsoft Research \\
  {\small \tt \{pshuang, xiaohe\}@microsoft.com}
\\\And 
  Li Deng\thanks{\hspace{1mm} Work performed when the author was at Microsoft Research.} \\
  Citadel Securities, LLC \\
  {\small \tt l.deng@ieee.org}}

\date{}

\begin{document}

\maketitle

\begin{abstract}
%TODO rewrite
We develop a technique for transfer learning in machine comprehension (MC) using a novel two-stage synthesis network (SynNet). Given a high-performing MC model in one domain, our technique aims to answer questions about documents in another domain, where we use no labeled data of question-answer pairs. Using the proposed SynNet with a pretrained model from the SQuAD dataset on the challenging NewsQA dataset, we achieve an F1 measure of 44.3\% with a single model and 46.6\% with an ensemble, approaching performance of in-domain models (F1 measure of 50.0\%) and outperforming the out-of-domain baseline of 7.6\%, without use of provided annotations.\footnote{\small Code will be available at \url{https://github.com/davidgolub/QuestionGeneration}}
\end{abstract}
\begin{figure}[t]
%\floatbox[{\capbeside\thisfloatsetup{capbesideposition={right,top},capbesidewidth=7cm}}]{figure}[\FBwidth]
{\includegraphics[width=1.1\textwidth]{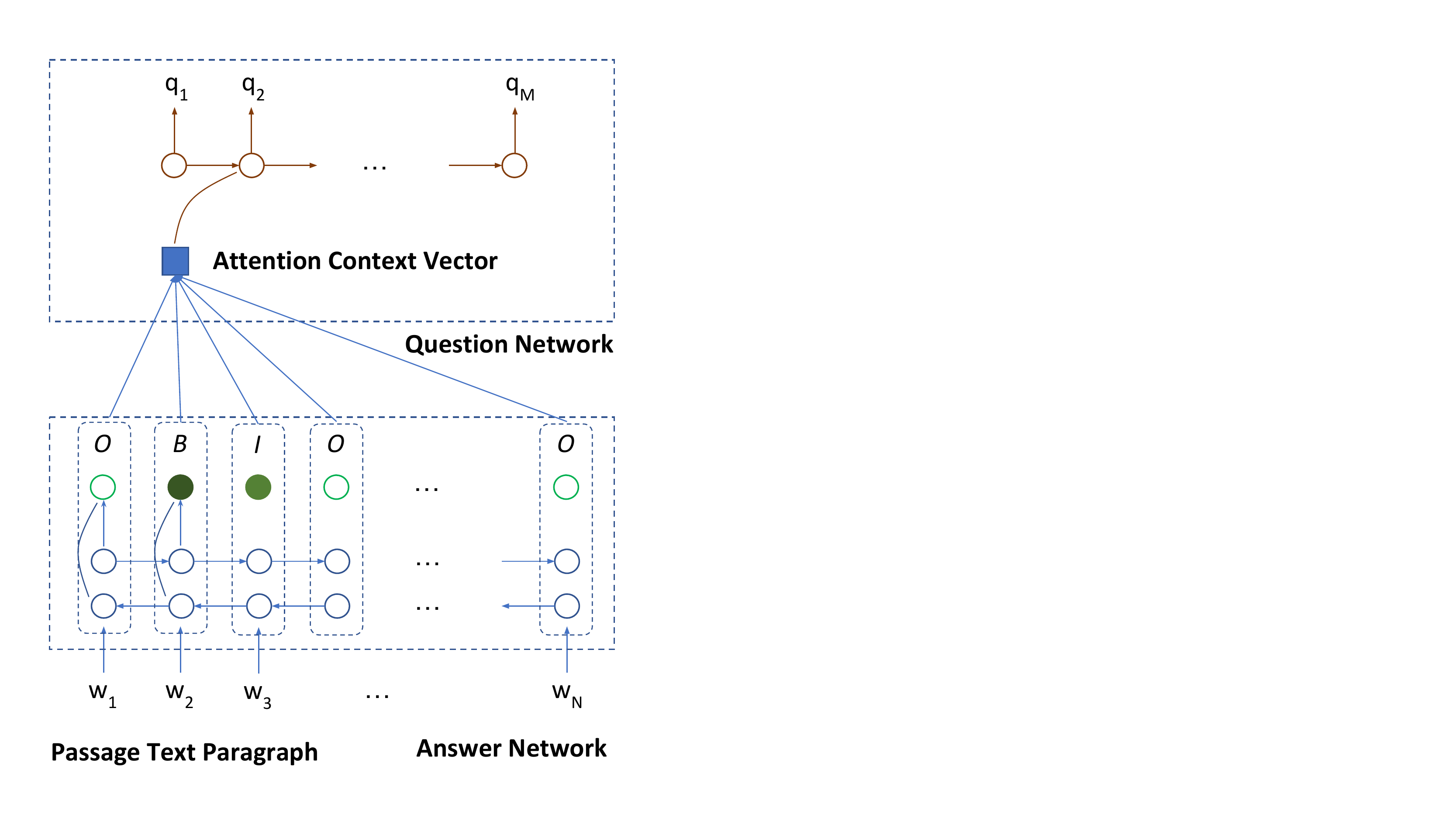}}
\vspace*{-12mm}
\caption{\small{Illustration of the two-stage SynNet. The SynNet is trained to synthesize the answer and the question, given the paragraph. The first stage of the model, an answer synthesis module, uses a bi-directional LSTM to predict IOB tags on the input paragraph, which mark out key semantic concepts that are likely answers. The second stage, a question synthesis module, uses a uni-directional LSTM to generate the question, while attending on embeddings of the words in the paragraph and IOB ids. Although multiple spans in the paragraph could be identified as potential answers, we pick one span when generating the question.
}\label{fig:test}}
\end{figure}

\section{Introduction}
%a.	Machine reading comprehension is really hot, have a lot of usage, progress very well. Lots of dataset proposed.
Machine comprehension (MC), the ability to answer questions over a provided context paragraph, is a key task in natural language processing. The rise of high-quality, large-scale human-annotated datasets for this task \cite{SQuAD, newsqa} has allowed for the training of data-intensive but expressive models such as deep neural networks \cite{multi, dynamic, BidirectionalAF}. Moreover, these datasets have the attractive quality that the answer is a short snippet of text within the paragraph, which narrows the search space of possible answer spans.

%b.	However, how to generalize is understudied. MC model learned from one domain to domain.
However, many of these models rely on large amounts of human-labeled data for training. Yet data collection is a time-consuming and expensive task. Moreover, direct application of a MC model trained on one domain to answer questions over paragraphs from another domain may suffer performance degradation. 

While understudied, the ability to transfer a MC model to multiple domains is of great practical importance. For instance, the ability to quickly use a MC model trained on Wikipedia to bootstrap a question-answering system over customer support manuals or news articles, where there is no labeled data, can unlock a great number of practical applications.

In this paper, we address this problem in MC through a two-stage synthesis network (SynNet). The SynNet generates synthetic question-answer pairs over paragraphs in a new domain that are then used in place of human-generated annotations to finetune a MC model trained on the original domain.

The idea of generating synthetic data to augment insufficient training data has been explored before. For example, for the target task of translation, \newcite{sennrich-haddow-birch:2016:P16-11} present a method to generate synthetic translations given real sentences to refine an existing machine translation system. 

However, unlike machine translation, for tasks like MC, we need to synthesize both the question and answers given the context paragraph. Moreover, while the question is a syntactically fluent natural language sentence, the answer is mostly a salient semantic concept in the paragraph, e.g., a named entity, an action, or a number, which is often a single word or short phrase.\footnote{This assumption holds for MC datasets such as SQuAD and NewsQA, but there are exceptions in certain subdomains of MSMARCO.} Since the answer has a very different linguistic structure compared to the question, it may be more appropriate to view answers and questions as two different types of data. Hence, the synthesis of a (question, answer) tuple is needed.

%In addition to the two-stage SynNet, we propose a training algorithm that consists of interleaving mini-batches of annotated-data and synthetic data as a form of regularization. Using our SynNet and training procedure enables us to significantly outperform directly applying an out-of-domain model.
In our approach, we decompose the process of generating question-answer pairs into two steps, answer generation conditioned on the paragraph, and question generation conditioned on the paragraph and answer. We generate the answer first because answers are usually key semantic concepts, while questions can be viewed as a full sentence composed to inquire the concept.

Using the proposed SynNet, we are able to outperform a strong baseline of directly applying a high-performing MC model trained on another domain. For example, when we apply our algorithm using a pretrained model on the Stanford Question-Answering Dataset (SQuAD)  \cite{SQuAD}, which consists of Wikipedia articles, to answer questions on the NewsQA dataset \cite{newsqa}, which consists of CNN/Daily Mail articles, we improve the performance of the single-model SQuAD baseline from 39.0\% to 44.3\% F1, and boost results further with an ensemble to 46.6\% F1, approaching results of previously published work of \newcite{newsqa} (50.0\% F1), without use of labeled data in the new domain. Moreover, an error analysis reveals that we achieve higher accuracy over the baseline on all common question types.

\section{Related Work}
\subsection{Question Answering}
Question answering is an active area in natural language processing with ongoing research in many directions \cite{semantic, hill2015goldilocks, golub2016character, chen2016thorough, hermann2015teaching}. Machine comprehension, a form of extractive question answering where the answer is a snippet or multiple snippets of text within a context paragraph, has recently attracted a lot of attention in the community. The  rise of large-scale human annotated datasets with over 100,000 realistic question-answer pairs such as SQuAD \cite{SQuAD}, NewsQA \cite{newsqa}, and MSMARCO \cite{msmarco}, has led to a large number of successful deep learning models \cite{lee2016learning, BidirectionalAF, dynamic, gated, wang2016machine}. 

\subsection{Semi-Supervised Learning}
Semi-supervised learning  has a long history (c.f. \newcite{chapelle2009semi} for an overview), and has been applied to many tasks in natural language processing such as dependency parsing \cite{koo2008simple}, sentiment analysis \cite{yang2015lcct},machine translation \cite{sennrich-haddow-birch:2016:P16-11}, and semantic parsing \cite{berant2014semantic,wang2015building,jia2016data}. Recent work generated synthetic annotations on unsupervised data to boost the performance of both reading comprehension and visual question answering models \cite{semi, ren2015exploring}, but on domains with some form of annotated data. There has also been work on generating high-quality questions \cite{DBLP:journals/corr/YuanWGSBSZT17, serban2016generating, labutov2015deep}, but not how to best use them to train a model. In contrast, we use the two-stage SynNet to generate data tuples to directly boost performance of a model on a domain with no annotations. 

\subsection{Transfer Learning}
Transfer learning \cite{pan2010survey} has been successfully applied to numerous domains in machine learning, such as machine translation \cite{zoph2016transfer}, computer vision, \cite{sharif2014cnn}, and speech recognition \cite{doulaty2015data}. Specifically, object recognition models trained on the large-scale ImageNet challenge \cite{russakovsky2015imagenet} have proven to be excellent feature extractors for diverse tasks such as image captioning (i.e., \newcite{lu2016knowing,fang2015captions, karpathy2015deep}) and visual question answering (i.e., \newcite{zhou2015simple, xu2016ask, fukui2016multimodal, yang2016stacked}), among others. In a similar fashion, we use a model pretrained on the SQuAD dataset as a generic feature extractor to bootstrap a QA system on NewsQA. 
%\textbf{Generative Adversarial Networks (GANs)}

%GANs \cite{goodfellow2014generative}, where optimization is framed as a minimax game between a generator and a discriminator, have been successfully applied to diverse tasks such as image generation \cite{deepgenadversarial}, text generation \cite{hu2017controllable}, and reading comprehension \cite{semi}. Similar to previous work, we use a generator model to bootstrap a discriminator. However, we find training the generator with simple maximum-likelihood estimation (MLE) is sufficient to bootstrap a strong reading comprehension model.

\section{The Transfer Learning Task for MC}
We formalize the task of machine comprehension below. Our MC model takes as input a tokenized question $q=\{q_0,q_1,...q_n\}$, a context paragraph $p=\{p_0,p_1,...p_n\}$, where $q_i, p_i$ are words, and learns a function $f(p,q) \mapsto \{a_{start},a_{end}\}$ where $a_{start}$ and $a_{end}$ are pointer indices into paragraph $p$, i.e., the answer $a=p_{a_{start}}...p_{a_{end}}$.

Given a collection of labeled paragraph, question, answer triples ${\{p, q, a\}}_{i=1}^{n}$ from a particular domain $s$, i.e., Wikipedia articles, we can learn a MC model $f_s(p, q)$ that is able to answer questions in that domain. 

However, when applying the model trained in one domain to answer questions in another, the performance may degrade. On the other hand, labeling data to train a model in the new domain is expensive and time-consuming. 

In this paper, we propose the task of transferring a MC system $f_s(p,q)$ that is trained in a source domain to answer questions over another target domain, $t$. In the target domain $t$, we are given an unlabeled set $p_t=\{p\}_{i=1}^{k}$ of $k$ paragraphs. During test time, we are given an unseen set of paragraphs, ${p^*}$, in the target domain, over which we would like to answer questions.

\section{The Model}
\subsection{Two-Stage SynNet}
To bootstrap our model $f_s$ we use a SynNet (Figure 1), which consists of answer synthesis and question synthesis modules, to generate data on $p_t$. 
Our SynNet learns the conditional probability of generating answer $a=\{a_{start}, a_{end}\}$ and question $q=\{q_1,...q_n\}$ given paragraph $p$, $P(q,a|p)$. We decompose the joint probability distribution $P(q,a|p)$ into a conditional probability distribution $P(q|p,a)P(a|p)$, where we first generate the answer $a$, followed by generating the question $q$ conditioned on the answer and paragraph. 

\subsubsection{Answer Synthesis Module}  

In our answer synthesis module we train a simple IOB tagger to predict whether each word in the paragraph is part of an answer or not. 

More formally, given a set of words in a paragraph $p=\{p_1...p_n\}$, our IOB tagging model learns the conditional probability of labels $y_1...y_n$, where $y_1\in{\text{IOB\textsubscript{START}}, \text{IOB\textsubscript{MID}}, \text{IOB\textsubscript{END}}}$ if a word $p_i$ is marked as an answer by the annotator in our train set, $\text{NONE}$ otherwise.  

We use a bi-directional Long-Short Term Memory Network (Bi-LSTM) \cite{hochreiter1997long} for tagging. Specifically, we project each word $p_i\mapsto p_i^*$ into a continuous vector space via pretrained GloVe embeddings \cite{pennington2014glove}. We then run a Bi-LSTM over the word embeddings $p_1^*,...p_n^*$ to produce a context-dependent word representation $h_1,...h_n$, which we feed into two fully connected layers followed by a softmax to produce our tag likelihoods for each word. 

We select all consecutive spans where $y\neq 
\text{NONE}$ produced by the tagger as our candidate answer chunks, which we feed into our question synthesis module for question generation.

\subsubsection{Question Synthesis Module} 
 Our question synthesis module learns the conditional probability of generating question $q=\{q_1,...q_n\}$ given answer $a={a_{start}, a_{end}}$ and paragraph $p={p_1...p_n}$, $P(q_1,...q_n|p_1...p_n,a_{start}, a_{end})$. We decompose the joint probability distribution of generating all the question words $q_1,...q_n$ into generating the question one word at a time, i.e. $\prod_{i=1}^{n}P(q_i|p,a,q_{1...i-1})$. 

The model is similar to an encoder-decoder network with attention \cite{bahdanau2014neural}, which computes the conditional probability $P(q_i|p_1...p_n,a_{start},a_{end},q_{1...i-1})$. We run a Bi-LSTM over the paragraph to produce context-dependent word representations $h=\{h_1,...h_n\}$. To model where the answer is in the paragraph, similar to \newcite{semi}, we insert answer information by appending a zero/one feature to the paragraph word embeddings. Then, at each time step $i$, a decoder network attends to both $h$ and the previously generated question token $q_{i-1}$ to produce a hidden representation $r_{i}$. Since paragraphs may often have named entities and rare words not present during training, we incorporate a copy mechanism into our models \cite{gu2016incorporating}.

We use an architecture motivated by latent predictor networks \cite{ling2016latent} to force the model to learn when to copy vs. directly predict the word, without direct supervision of what action to choose. Specifically, at every time step $i$, two latent predictors generate the probability of generating word $w_i$, a pointer network $C_p$ \cite{pointer} which can copy a word from the context paragraph, and a vocabulary predictor $V_p$ which directly generates a probability distribution of choosing a word $w_i$ from a predefined vocabulary. The likelihood of choosing predictor $k$ at time step $i$ is proportional to $w_k  r_i$, and the likelihood of predicting question token $q_i$ is given by $q_i^*=p^{v}l^{v}(w_i) + (1-p^{v})l^{c}(w_i)$, where $v$ represents the vocabulary predictor and $c$ represents the copy predictor, and $l(w_i)$ is the likelihood of the word given by the predictor.\footnote{Since we only have two predictors, $p^{c} = 1 - p^{v}$} For training, since no direct supervision is given as to which predictor to choose, we minimize the cross entropy loss of producing the correct question tokens $\sum_{j=1}^{n}{-log(q_j^*)}$ by marginalizing out latent variables using a variant of the forward-backward algorithm (see \newcite{ling2016latent} for full details).

During inference, to generate a question $q_1...q_n$, we use greedy decoding in the following manner. At time step $i$, we select the most likely predictor ($C_p$ or $V_p$), followed by the most likely word $q_i$ given the predictor. We feed the predicted word as input at the next timestep back into the decoder until we predict the end symbol, $\text{END}$, after which we stop decoding.

\begin{table*}[tp]
{\small
%{\large
		
		\label{newsqa_questions}

\begin{center}
\begin{tabular}{ | p{21em} | p{21em} | } 
\hline 
Snippet of context paragraph (answer in bold) & Generated questions (bold) vs. human questions \\
\hline 
 ...At this point, some of these used-luxe models have been around so long that they almost qualify as vintage throwback editions. Recently, \textbf{Consumer Report } magazine issued its list of best and worst used cars, and divvied them up by price range ... &
\textbf{What magazine made best used cars in the USAF?}
\newline
Who released a list of best and worst used cars
\\
\hline
...A	high	court	in	northern	India	on	Friday	acquitted	a	wealthy	businessman	facing	the	death	sentence	for	the	killing	of	a	teen	in	a	case	dubbed "the	house of horrors.``	Moninder	Singh	Pandher	was	sentenced	to	death	by	a	lower	court	in	February.	The	teen	was	\textbf{one	of	19}	victims	--	children	and ...& \textbf{How many victims were in India ?}  \newline What	was	the	amount	of	children	murdered	? \\ 
\hline
Joe Pantoliano has met \textbf{ with the Obama and McCain camps} to promote mental health and recovery. Pantoliano, founder and president of the eight-month-old advocacy organization No Kidding, Me Too, released a teaser of his new film about various forms of mental illness... 
& \textbf{Which two groups did Joe Pantoliano meet with?} \newline
Who did he meet with to discuss the issue?
\\
\hline
...Former boxing champion Vernon Forrest , 38 , was shot and killed in \textbf{southwest Atlanta , Georgia} , on July 25 . A grand jury indicted the three suspects -- Charman Sinkfield , 30 ; Demario Ware , 20 ; and Jquante...
& \textbf{Where was the first person to be shot ?}
\newline Where	was	Forrest	killed?
\\ 
\hline

\end{tabular}
\end{center}
\caption{Randomly sampled paragraphs and corresponding synthetic vs. human questions from the NewsQA train set. Human-selected answers from the train set were used as input.}
}
 
\end{table*}

\subsection{Machine Comprehension Model}
Our machine comprehension model $f(p,q)\mapsto a$ learns the conditional likelihood of predicting answer pointers $a=\{a_{start}, a_{end}\}$ given paragraph $p$ and question $q$, $P(a|p,q)$. In our experiments we use the open-source Bi-directional Attention Flow (BiDAF) network \cite{BidirectionalAF}\footnote{See https://github.com/allenai/bi-att-flow} since it is one of the best-performing models on the SQuAD dataset,\footnote{See https://rajpurkar.github.io/SQuAD-explorer/ for latest results} although we note that our algorithm for data synthesis can be used with any MC model.  

\subsection{Algorithm Overview}
Having given an overview of our SynNet and a brief overview of the MC model we describe our training procedure, which is illustrated in Algorithm \ref{alg:inference_process}.
\begin{algorithm}[!t]

	\SetKwInOut{Input}{Input}
	\SetKwInOut{Output}{Output}
	\Input{$x_s={\{p_s, q_s, a_s\}}_{i=1}^{n}$ triplets from source domain $s$; pretrained MC model on $s$, $f_s(p,q)\mapsto \{a_{start}, a_{end}\}$; paragraphs from target domain $t$, ${p}_{j=1}^{m}$}
	\Output{MC model on target domain, $f_t(p, q) \mapsto \{a_{start}, a_{end}\}$}
	Train SynNet $g$ to maximize $P(q,a|p)$ on source $s$\;
	Generate samples $x_t=(q,a|p)_{i=1}^{k}$ on text in target domain $t$\;
	Use $x_s \cup x_t$ to finetune MC model $f_s$ on domain $t$. For every batch sampled from $x_t$, sample $k$ batches from $x_s$\;
	\caption{Training Algorithm}
	\label{alg:inference_process}
\end{algorithm}
\subsection{Training} 
Our approach for transfer learning consists of several training steps. 
First, given a series of labeled examples $x_s={\{p_s, q_s, a_s\}}_{i=1}^{n}$ from domain $s$, paragraphs ${p}_{j=1}^{m}$ from domain $t$, and pretrained MC model $f_s(p,q)$, we train the SynNet $g_s$ to maximize the likelihood of the question-answer pairs in $s$.

Second, we fix our SynNet $g_s$ and we sample $x_t={\{p_t, q_t, a_t\}}_{i=1}^{k}$ question-answer pairs on the paragraphs in domain $t$. Several examples of generated questions can be found in Table 1.

We then transfer the MC model originally learned on the source domain to the target domain $t$ using SGD on the synthetic data. However, since the synthetic data is usually noisy, we alternatively train the MC model with mini-batches from $x_s$ and $x_t$, which we call \textit{data-regularization}. Every $k$ batches from $x$, we sample 1 batch of synthetic data from $x'$, where $k$ is a hyper-parameter, which we set to 4. Letting the model encounter many examples from source domain $s$ serves to regularize the distribution of the synthetic data in the target domain with real data from $s$. We checkpoint finetuned model $f^*_s$ every $i$ mini-batches, $i=1000$ in our experiments, and save a copy of the model at each checkpoint. 

At test time, to generate an answer, we feed paragraph $p=\{p_0,p_1,...p_n\}$ and question $q$ through our finetuned MC model $f^*(p,q)$ to get $P(p_i=a_{start})$, $P(p_i=a_{end})$ for all $i\in{1...n}$. We then use dynamic programming \cite{BidirectionalAF} to find the optimal answer span $\{a_{start}, a_{end}\}$. To improve the stability of using our model for inference, we average the predicted answer likelihoods from model copies at different checkpoints, which we call $checkpoint-averaging$, prior to running the dynamic programming algorithm.

\section{Experimental Setup}
We summarize the datasets we use in our experiments, parameters for our model architectures, and training details.

The SQuAD dataset consists of approximately 100,000 question-answer pairs on Wikipedia, 87,600 of which are used for training, 10,570 for development, and an unknown number in a hidden test set. The NewsQA dataset consists of 92,549 train, 5,166 development and 5,165 test questions on CNN/Daily Mail news articles. Both the domain type (i.e., news) and question types differ between the two datasets. For example, an analysis of a randomly generated sample of 1,000 questions from both NewsQA and SQuAD \cite{newsqa} reveals that approximately 74.1\% of questions in SQuAD require word matching or paraphrasing to retrieve the answer, as opposed to 59.7\% in NewsQA. As our test metrics, we report two numbers, exact match (EM) and F1 score.

We train a BIDAF model on the SQuAD train dataset and use a two-stage SynNet to finetune it on the NewsQA train dataset. 

We initialize word-embeddings for the BIDAF model, answer synthesis module, and question synthesis module with 300-dimensional-GloVe vectors \cite{pennington2014glove} trained on the 840 Billion Words Common Crawl corpus. We set all embeddings of unknown word tokens to zero. 

For both the answer synthesis and question synthesis module, we use a vocabulary of size 110,179. We use LSTMs with hidden states of size 150 for the answer module vs. those of size 100 for the question module since the answer module is less  memory intensive than the question module.

We train both the answer and question module with Adam \cite{kingma2014adam} and a learning rate of 1e-2. We train a BIDAF model with the default hyperparameters provided in the open-source repository. To stop training of the question synthesis module, after each epoch, we monitor both the loss as well as the quality of questions generated on the SQuAD development set. To stop training of the answer synthesis module, we similarly monitor predictions on the SQuAD development set.

\begin{table*}[htp]
\begin{center}
\begin{tabularx}{1.0\textwidth}{ l l c c c c c} \\
 \toprule
 Method & System & EM & F1 \\
\midrule 
Transfer Learning  & $M_{sq}$ (SQuAD baseline) & 24.9 & 39.0 \\
 & $M_{sq}$ + $A_{gen}$ + $Q_{gen}$ (single model on NewsQA) &  26.6 & 40.9 \\
 & $M_{sq}$ + $A_{gen}$ + $Q_{gen} + M^*_{sq}$ (single model on NewsQA) &  29.0 & 43.1 \\
 & $M_{sq}$ + $A_{gen}$ + $Q_{gen}$ (single model on NewsQA, cpavg) &  30.6 & 44.3 \\
 & $M_{sq}$ + $A_{gen}$ + $A_{ner}$ + $Q_{gen}$ (4-model ensemble, cpavg) & 32.8 & 46.6 \\
 & $M_{sq}$ + $A_{gen}$ + $A_{ner}$ + $Q_{gen}$ + $M^*_{sq}$ (4-model ensemble, cpavg) & \textbf{33.0} & \textbf{46.6} \\
\midrule
Supervised Learning & Barb-LSTM on NewsQA \cite{newsqa} & 34.9 & 50.0  \\
 & Match-LSTM on NewsQA \cite{newsqa} & 34.1 & 48.2 \\
 & BIDAF on NewsQA & 37.1 & 52.3 \\ 
 & BIDAF on SQuAD finetuned on NewsQA & 37.3 & 52.2  \\ 
\hline
\end{tabularx}
\end{center}
\caption{\textbf{Main Results}. Exact match (EM) and span F1 scores on the NewsQA test set of a BIDAF model finetuned with our SynNet. $M_{sq}$ refers to a baseline BIDAF model trained on SQuAD, $A_{gen}$, $Q_{gen}$ refers to using answers generated from our SynNet respectively to finetune the model on NewsQA, $A_{ner}$ refers to using answers extracted from a standard NER system to generate questions, $cpavg$ refers to using checkpoint-averaging, and $M^*_{sq}$ refers to using the baseline SQuAD model in the ensemble.}

\label{tab:results}
\end{table*}

\begin{table}[ht]
%\begin{center}
\hspace*{3em}
\begin{tabularx}{0.32\textwidth}{ l c c} 
\\
\noalign{\noindent\rule{5cm}{0.5pt}} System & EM & F1 \\
\noalign{\noindent\rule{5cm}{0.5pt}}  $M_{newsqa}$ & 46.3 & 60.8 \\
$M_{newsqa}$ + $S_{net}$ & 47.9 & 61.5 \\
\noalign{\noindent\rule{5cm}{0.5pt}}
\end{tabularx}

%\end{center}
\caption{\textbf{NewsQA to SQuAD}. Exact match (EM) and span F1 results on SQuAD development set of a NewsQA BIDAF model baseline vs. one finetuned on SQuAD using the data generated by a 2-stage SynNet ($S_{net}$).}
\label{tab:results}  
\end{table}

To train the question synthesis module, we only use the questions provided in the SQuAD train set. However, to train the answer synthesis module, we further augment the human-annotated labels of each paragraph with tags from a simple NER system\footnote{https://spacy.io/} because labels of answers provided in the train set are underspecified, i.e., many words in the paragraph that could be potential answers are not labeled. Therefore, we assume any named entities could also be potential answers of certain questions, in addition to the answers explicitly labeled by annotators.

To generate question-answer pairs on the NewsQA train set using the SynNet, we first run every paragraph through our answer synthesis module. We then randomly sample up to 30 candidate answers extracted by our module, which we feed into the question synthesis module. This results in 250,000 synthetic question-answer pairs that we can use to finetune our MC model.

\begin{table}[ht]
\begin{center}
\begin{tabularx}{0.3\textwidth}{ l l l || l l c c} \\
\noalign{\noindent\rule{7.5cm}{0.5pt}}
 \textbf{A)} & EM & F1 & \textbf{B)} & EM & F1 \\
\noalign{\noindent\rule{7.5cm}{0.5pt}}
k=0 & 27.2 & 40.5 & 2s + $A_{ner}$ & 22.8 & 36.1 \\
k=2 & 29.8 & 43.9 & all + $A_{ner}$ & 27.2 & 40.5 \\
k=4 & 30.4 & 44.3 & 2s + $A_{oracle}$ & 31.3 & 45.2 \\ 
 & & & all + $A_{oracle}$ & 32.5 & 46.8 \\
\noalign{\noindent\rule{7.5cm}{0.5pt}}
\end{tabularx}
\end{center}
\caption{\textbf{Ablation Studies}. Exact match (EM) and span F1 results on NewsQA test set of a BIDAF model finetuned with a 2-stage SynNet. In study A, we vary $k$, the number of mini-batches from SQuAD for every batch in NewsQA. In study B, we set $k=0$, and vary the answer type and how much of the paragraph we use for question synthesis. $2-sent$ refers to using two sentences before answer span, while $all$ refers to using the entire paragraph. $A_{ner}$ refers to using an NER system and $A_{or}$ refers to using the human-annotated answers to generate questions.}
\label{tab:results}
\vspace*{-3mm}
\end{table}
\section{Experimental Results}

We report the main results on the NewsQA test set (Table 2), report brief results on SQuAD (Table 3), conduct ablation studies (Table 4), and conduct an error analysis.

\subsection{Results}
We compare to the best previously published work, which trains BARB \cite{newsqa} and Match-LSTM \cite{wang2016machine}  architectures, and a BIDAF model we train on NewsQA. Directly applying a BIDAF model trained on SQuAD to predict on NewsQA leads to poor performance with an F1 measure of 39.0\%, 13.2\% lower than one trained on labeled NewsQA data. Using the 2-stage SynNet already leads to a slight boost in performance (F1 measure of 40.9\%), which implies that having exposure to the new domain via question-answer pairs provides important signal for the model during training. With checkpoint-averaging, we see an additional improvement of 3.4\% (F1 measure of 44.3\%). When we ensemble a BIDAF model trained on questions and answers from the SynNet with three BIDAF models trained on questions by $Q_{gen}$ and answers from a generic NER system, we have an additional 2.3\% performance boost. Finally, when we ensemble the original BIDAF model trained on SQuAD in the ensemble, we boost the EM further by 0.2\%. Our final system achieves an F1 measure of 46.6\%, approaching previously published results of 50.0\%. The results demonstrate that using the proposed architecture and training procedure, we can transfer a MC model from one domain to another, without use of annotated data.

We also evaluate the SynNet on the NewsQA-to-SQuAD direction. We directly apply the best setting from the other direction and report the result in Table 3. The SynNet improves over the baseline by 1.6\% in EM and 0.7\% in F1. Limited by space, we leave out ablation studies in this direction.

\subsection{Ablation Studies}
To better understand how various components in our training procedure and model impact overall performance we conduct several ablation studies, as summarized in Table 4. 
\subsubsection{Answer Synthesis}
We experiment with using the answer chunks given in the train set, $A_{oracle}$, to generate synthetic questions, versus those from an NER system, $A_{ner}$. Results in Table 4(A) show that using human-annotated answers to generate questions leads to a significant performance boost over using answers from an answer generation module. This supports the hypothesis that the answers humans choose to generate questions for provide important linguistic cues for finetuning the machine comprehension model. 
\subsubsection{Question Synthesis}
To see how copying impacts performance, we explore using the entire paragraph to generate the question vs. only the two sentences before and one sentence after the answer span and report results in Table 4(B). On the NewsQA train set, synthetic questions that use 2 sentences contain an average of 3.0 context words within 10 words to the left and right of the answer chunk, those that use the entire context have 2.1 context words, and human generated questions only have 1.7 words. Training with generated questions that have a large amount of overlap with words close to the answer span (i.e., those that use 2-sentences vs. entire context for generation) leads to models that perform worse, especially with synthetic answer spans and no data regularization (35.6\% F1 vs. 34.3\% F1). One possible reason is that, according to analysis in \newcite{newsqa}, significantly more questions in the NewsQA dataset require paraphrase, inference, and synthesis as opposed to word-matching.

\subsubsection{Model Finetuning}
To see how the quantity of synthetic questions encountered during training impacts performance, we use $k=\{0, 2, 4\}$ mini-batches from SQuAD for every synthetic mini-batch from NewsQA to finetune our model, and average the prediction of 4 checkpointed models during testing. As we see from the results, letting the model to encounter data from human annotations, although from another domain, serves as a key form of data-regularization, yielding consistent improvement as $k$ increases. We hypothesize this is because the data distribution of machine-generated questions is different than human-annotated ones; our batching scheme provides a simple way to prevent over-fitting to this distribution.

\subsection{Error Analysis} 
In this section we provide a qualitative analysis of some of our components to help guide further research in this task.
\subsubsection{Answer Synthesis}
We randomly sample and present a paragraph with answers extracted by our answer synthesis module (Tables 5 and 6). Although the module appears to have high precision, i.e., it picks up entities such as the ``Atlantic Paranormal Society'', it misses clear entities such as ``David Schrader'', which suggests training a system with full NER/POS tags as labels would yield better results, and also explains why augmenting synthetic data generated by SynNet with such tags leads to improved performance.

\noindent\begin{table}[t!]
\small
\begin{tabularx}{0.5\textwidth}{l}\\
\toprule
They are ghost hunters , or , as they prefer to be called , \\ paranormal investigators . `` Ghost-Hunters '', which airs a \\ special live show at \textbf{ 7 p.m. Halloween night , } is helping lift \\ the stigma once attached to paranormal investigators . The \\ show has become so popular that the group featured in each \\ episode -- \textbf{ Atlantic Paranormal Society } - has spawned \\ imitators across \textbf{ United States } and affiliates in \textbf{ countries } . \\ TAPS , as the ``\textbf{ Hunters'' } group is informally known , even \\ has its own ``\textbf{ Reality Radio''} show , magazine , lecture tours , \\ T-shirts -- and groupies . ``\textbf{ Hunters''} has made creepy cool , \\ says David Schrader , a paranormal investigator and co-host of \\ ``\textbf{ Radio }'', a radio show that investigates paranormal activity.\\
\bottomrule
%\hline
\end{tabularx}

\caption{Sample predictions from our answer synthesis module.}
\end{table}

\subsubsection{Question Synthesis}
We randomly sample synthetic questions generated by our module and present our results in Table 6. Due to the copy mechanism, our module has the tendency to directly use many words from the paragraph, especially common entities, such as ``Oklahoma'' in the example. Thus, one way to generate higher-quality questions may be to introduce a cost function that promotes diversity during decoding, especially within a single paragraph. In turn, this would expose the RC model to a larger variety of training examples in the new domain, which can lead to better performance.

\begin{table}[t]
\small
\begin{tabularx}{0.5\textwidth}{l}% \\
\toprule
What is Oklahoma's unemployment rate until Oklahoma City ? \\ 
What was the manager of the Oklahoma City agency ? \\
How many companies are in Oklahoma City ? \\
How many workers may Oklahoma have as fair hold ? \\
Who said the bureau has already hired civilians to choose \\
What was the average hour manager of Oklahoma City ? \\
How much would Oklahoma have a year to be held \\
What year did Oklahoma 's census build job industry ?\\
\bottomrule
\end{tabularx}
\caption{
Predictions from the question synthesis module on a subset of a paragraph.}
\end{table}

\begin{figure}[t!]
    \vspace*{-11mm}
    \includegraphics[width=0.50\textwidth]{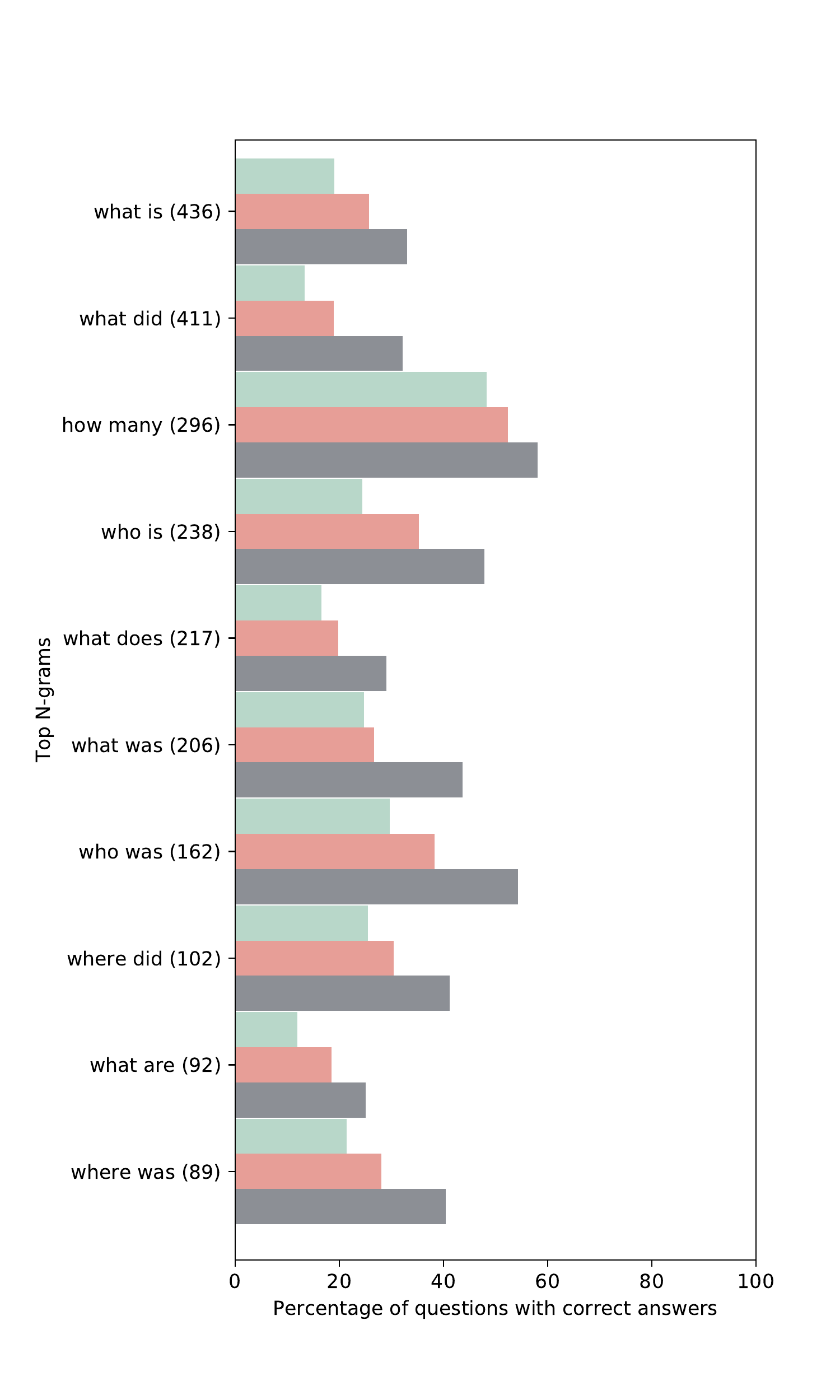}
    \vspace*{-10mm}
      \caption{NewsQA accuracy of baseline BIDAF model trained on SQuAD (light green), vs. model finetuned with our method (red) vs. one trained from scratch on NewsQA (dark grey).}
\end{figure}

\subsubsection{Machine Comprehension Model}
 We examine the performance over various question types of a finetuned BIDAF on NewsQA vs. one trained on NewsQA vs. one trained on SQuAD (Figure 2). Finetuning with SynNet improves performance over all question types given, with the largest performance boost on location and person-identification questions. Similarly, models trained on synthetic questions tend to approach in-domain performance on numeric and person-identification questions, but still struggle with questions that require higher-order reasoning, i.e. those starting with ``what was'' or ``what did''. Designing a question generator that explicitly requires such reasoning may be one way to further bridge the gap in performance.

\section{Conclusion}
We introduce a two-stage SynNet for the task of transfer learning for machine comprehension, a task which is both challenging and of practical importance. With our network and a simple training algorithm where we generate synthetic question-answer pairs on the target domain, we are able to generalize a MC model from one domain to another with no annotated data. We present strong results on the NewsQA test set, with a single model improving performance of a baseline BIDAF model by 5.3\% and an ensemble by 7.6\% F1. Through ablation studies and error analysis, we provide insights into our methodology on the SynNet and MC models that can help guide further research in this task. 

\section*{Acknowledgments}
We would like to thank Yejin Choi and Luke Zettlemoyer for helpful discussions concerning this work.
%Do not number the acknowledgment section.

\bibliography{emnlp2017}
\bibliographystyle{emnlp_natbib}

\end{document}